\def\year{2021}\relax
\documentclass[letterpaper]{article} 
\usepackage{aaai21}  
\usepackage{times}  
\usepackage{helvet} 
\usepackage{courier}  
\usepackage[hyphens]{url}  
\usepackage{graphicx} 
\usepackage{natbib}  
\usepackage{caption} 

\usepackage{subfigure}
\usepackage{multirow}
\usepackage{makecell}
\usepackage{booktabs}

\usepackage{algorithm}
\usepackage{algorithmic}
\usepackage{amsmath}
\usepackage{amsthm}
\usepackage{ifthen}
\usepackage[switch]{lineno}
\title{Show Me How To Revise:\\ Improving Lexically Constrained Sentence Generation with XLNet}
\author {
        Xingwei He, 
        Victor O.K. Li \\
}
\affiliations {
    Department of Electrical and Electronic Engineering, The University of Hong Kong, 
    Hong Kong, China \\
    hexingwei15@gmail.com, vli@eee.hku.hk
}

\begin{document}

\maketitle

\begin{abstract}
  Lexically constrained sentence generation allows the incorporation of 
  prior knowledge such as lexical constraints into the output. This technique has been applied to 
  machine translation, and dialog response generation. 
  Previous work usually used Markov Chain Monte Carlo (MCMC) sampling to generate lexically constrained sentences, 
  but they randomly determined the position to be edited and the action to be taken, 
  resulting in many invalid refinements. 
  To overcome this challenge, we used a classifier to instruct the MCMC-based models where and how to refine the candidate sentences. 
  First, we developed two methods to create synthetic data on which the pre-trained model 
  is fine-tuned to obtain a reliable classifier. 
  Next, we proposed a two-step approach, ``Predict and Revise", for constrained sentence generation. 
  During the \textit{predict} step, we leveraged the classifier to 
  compute the learned prior for the candidate sentence. 
  During the \textit{revise} step, we 
  resorted to MCMC sampling to revise the candidate sentence by conducting 
 a sampled action at a sampled position drawn from the learned prior. 
  We compared our proposed models with many strong baselines on two tasks, generating sentences with lexical constraints and text infilling. 
  Experimental results have demonstrated that our proposed model performs much better than the previous work in terms of 
  sentence fluency and diversity. 
  Our code and pre-trained models are available at \url{https://github.com/NLPCode/MCMCXLNet}. 
  \end{abstract}

\section{Introduction}

Recently, there has been much interest in generating sentences in a controlled manner.
Lexically constrained sentence generation aims to incorporate some pre-specified keywords or phrases 
into the generated sentences, and has been widely used for many natural language processing (NLP) 
tasks. For example, Mou et al. \shortcite{mou2016sequence} alleviated generating universal dialog responses 
by injecting a keyword into the dialogue replies. 
Some researchers \cite{Hokamp2017LexicallyCD,post-vilar-2018-fast} 
incorporated the domain-specific terminologies into the translated sentences.

To generate sentences with lexical constraints, Mou et al. \shortcite{mou2015backward} 
proposed a novel backward and forward language model (B/F-LM). 
Liu et al. \shortcite{Liu2019BFGANBA} extended B/F-LM by introducing a discriminator. 
However, the capability of these models is limited, as they can generate sentences
with only one lexical constraint. 
To generate sentences with multiple lexical constraints, 
Hokamp and Liu \shortcite{Hokamp2017LexicallyCD} proposed grid beam search (GBS) by controlling the decoding process. 
But this may degrade the quality of the generated sentence 
 since the constraints are not considered when generating previous tokens. 
Applying GBS to unconditional generative models will suffer from this issue more seriously 
because the solution space is much wider than that of conditional generative models. 
In addition, beam search-based models tend to generate generic and repetitive sentences 
in unconditional text generation \cite{Holtzman2020TheCC}.

\begin{figure*}[!t]
  \centering
  \includegraphics[width=0.9\textwidth]{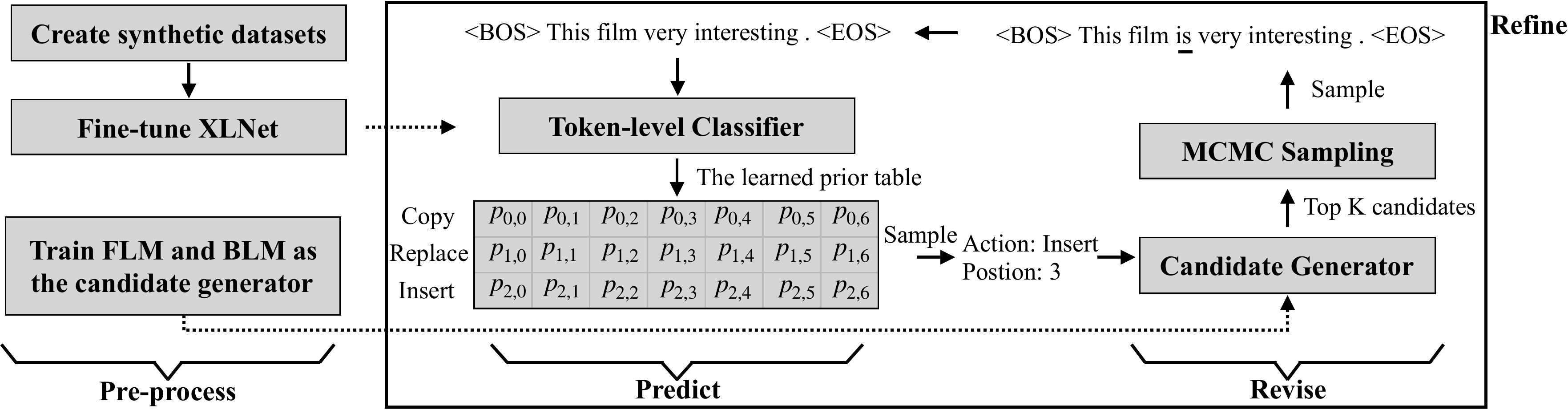}
  \caption{Illustration of our approach. In the \textit{predict} step, we leveraged the token-level classifier to 
  compute the learned prior for the current sentence. 
  Then, we sampled a proposed action and position from the learned prior. 
   In the \textit{revise} step, we generated the top $K$ candidate sentences, 
  from which we sampled a candidate sentence with MCMC.}
  \label{example}
\end{figure*}

Another line of work applies MCMC sampling to lexically constrained sentence generation. 
Berglund et al. \shortcite{berglund2015bidirectional} first used Gibbs sampling to generate sentences 
from the bidirectional RNN. 
Then, Wand and Cho \shortcite{wang-cho-2019-bert} extended Gibbs sampling to generate sentences from 
BERT \cite{Devlin2019BERTPO}. 
Su et al. \shortcite{su2018incorporating} incorporated soft constraints such as sentiments via discriminators
into the generated sentences. 
Furthermore, Miao et al. \shortcite{miao2019cgmh} proposed CGMH, which can generate 
flexible-length sentences under the given lexical constraints with 
replacement, insertion, and deletion actions. 
Different from GBS, 
MCMC-based models generate the constrained sentence via many refinements instead of one pass, 
which allows them to utilize both the past and future contexts to refine tokens one by one iteratively.

However, previous MCMC-based models typically conduct many invalid and redundant refinements because of the 
randomly chosen actions and positions. 
This problem is more serious when generating sentences with lexical constraints  
since it needs many more operations to generate a complete sentence. 
To demonstrate this problem, we showed an example in Figure \ref{example}. 
It is easy for us to realize that a token should be inserted before ``very''
to complete this sentence. Therefore, operations such as replacing a token with another token or inserting a token at some other position 
are very likely to be rejected since these refinements can hardly 
improve the quality of the candidate sentence. 
Without sufficient refinements, these models will risk generating ungrammatical sentences.

To reduce the redundant refinements conducted by MCMC-based models, 
we proposed a two-step approach, ``Predict and Revise", for constrained sentence generation, 
shown in Figure \ref{example}. 
Intuitively, suppose a classifier can tell the
model where and how to refine the candidate sentence, then we
will undoubtedly avoid many invalid operations, thus improving
the quality of the generated sentences. 
To train a reliable classifier, we proposed two methods (random and masked language model) 
to create synthetic data. 
When creating synthetic data for the replacement action, 
the random method replaces the chosen token with a randomly sampled token, 
while the masked language model method masks the chosen token and 
leverages the permutation-based language model, i.e., XLNet \cite{Yang2019XLNetGA}, to predict the masked token.  
A token from the top $N$ tokens is drawn as the replaced token. 
These two methods are complementary to each other, contributing to training a classifier. 
Then, we fine-tuned XLNet on the synthetic dataset to obtain a token-level classifier, 
which provides accurate revision information 
including the position to be revised and the action to be conducted. 
In the \textit{predict} step, an action and a position are drawn 
from the learned prior given by the classifier for the candidate sentence. 
In the \textit{revise} step, we used the candidate generator to generate the top $K$ candidate sentences, 
and drew a candidate sentence with MCMC. 

The main contributions of our work are threefold: 
(1) we proposed two approaches to create the synthetic dataset and 
fine-tuned XLNet to obtain a reliable classifier; 
(2) we proposed a two-step approach for constrained sentence generation, 
guided by the classifier to refine the candidate sentence; 
(3) we conducted extensive experiments on two tasks, generating sentences with lexical constraints and text infilling. 
Experiment results show that our proposed model outperforms previous baselines in sentence fluency and diversity.

\section{Problem Definition}
\subsection{Generating Sentences with Lexical Constraints} 
Generating sentences with lexical constraints aims to incorporate the given lexical constraints 
into the output. Given the constraints $c_1, c_2, \cdots, c_k$, this 
task is defined as follows:
\begin{eqnarray}\label{eq:4}
  X^* = \arg\max_{X} P(X|c_1, c_2, \cdots, c_k),
\end{eqnarray} 
where $X$ is the sentence containing the given lexical constraints. 
This task is meant to find a fluent sentence by maximizing the conditional probability. \\
\subsection{Text Infilling} 
Text infilling aims to infill the missing portions of a sentence based on the past and future contexts 
to make the sentence complete and fluent. 
In this paper, we follow a more general definition of text infilling
\cite{Zhu2019TextI}, where the number of missing portions is arbitrary, and each portion can be infilled with 
an arbitrary number of tokens. 
Given an incomplete sentence 
$X^B=\{x_1,\cdots,x_{i-1},\underline{m},x_{i+1},\cdots,x_{j-1},\underline{m},x_{j+1},\cdots,x_n\}$,
 an arbitrary number of tokens can be filled into each blank $\underline{m}$ to make it meaningful. 
Therefore, text infilling can be formulated as follows:
\begin{eqnarray}\label{eq:4_2}
  X^* = \arg\max_{X} P(X|X^B).
\end{eqnarray} 
Text infilling is a special case of constrained sentence generation, 
where the known portions can be regarded as lexical constraints. 
Compared with generating sentences with lexical constraints, 
tokens are permitted to be inserted only at specific positions, where blanks are located. 

\section{Methodology}
The overview of our proposed approach is demonstrated in Figure \ref{example}.
In this section, we will first introduce how to create the synthetic 
dataset for training a classifier. 
Then, we will describe how to train a classifier and use it to compute
the learned prior for the candidate sentence. 
Finally, we will demonstrate the process of generating candidate sentences with the candidate generator and 
refining the current candidate sentence with MCMC sampling. 

\subsection{Creating the Synthetic Dataset}\label{syn}
We aim to obtain a three-class token-level classifier, 
which is expected to tell us how to refine the candidate sentence. 
We use labels 0, 1, 2 to indicate copy, replacement, and insertion actions. 
The copy action indicates that the current token does not need to be modified. 
The replacement action indicates that the current token should be replaced with another token 
to make the sentence more fluent. 
Similarly, the insertion action means some tokens should 
be inserted before the current token to complete the sentence. 
To train a reliable classifier, we will create the synthetic dataset $D=\{(X, L)\}$, 
where $(X, L)$ denotes a data instance. 
We used One-Billion-Word corpus\footnote{http://www.statmt.org/lm-benchmark/} to construct the synthetic dataset. 
Each time we randomly selected a sentence from One-Billion-Word corpus. Assume the selected sentence is  
``Opponents of the tariff say U.S. manufacturing would suffer under the climate bill 
regardless of trade policy changes.” 
Then, we randomly chose a continuous segment from this sentence,  
“manufacturing would suffer under the climate bill.” 

To construct synthetic data for the insertion action, we randomly deleted some tokens from the selected part. 
Then, we obtained a synthetic input pair, $X$=\{$<$BOS$>$, manufacturing, suffer, under, the, climate, bill, $<$EOS$>$\} and 
$L=\{0,2,2,0,0,0,0,2\}$. 
In the above example, the labels for ``manufacturing” and ``$<$EOS$>$'' are 2, 
which means we need to insert something before these tokens. 
Since we deleted ``would'' from the selected sub-sentence, 
the label for ``suffer'' is also 2. Labels for the remaining tokens are 0. 
``$<$BOS$>$'' and ``$<$EOS$>$'' are special tokens, which represent the start and end of a sentence, respectively. 

We proposed two methods (random and masked language model) to create synthetic data for the replacement action. 
The random method replaces the chosen token with a randomly sampled token. For example, replacing 
the token ``would'' with ``(''. 
In this case, the classifier can easily infer the label for ``('' is 1.
Nevertheless, this case is far from the real mistakes made by humans. 
Therefore, the classifier will struggle to infer labels in scenarios 
where tense errors appear, or some inappropriate words are used. 
To solve this problem, the masked language model method uses XLNet to mimic mistakes made by humans. Specifically, 
we first replaced the chosen token, e.g., ``would'' with the masked token ``$<$MASK$>$''. 
Then, we used XLNet to predict the masked token ``$<$MASK$>$'' 
and sampled a token from the top $N$ predicted tokens with the highest probability. 
(In our experiments, we set $N$ to 20). 
Before using XLNet to create synthetic data, we fine-tuned it on 
the masked dataset constructed with One-Billion-Word 
to give reliably predicted tokens. 
These two methods are complementary to each other, and we combined them to create the synthetic dataset.

\subsection{The Token-level Classifier}\label{cla}
After getting the synthetic dataset, we fine-tuned XLNet on them to get a token-level classifier. 
We chose XLNet as the classifier because it can not only be used as a classifier 
but also serve as a transformer-based language model \cite{Vaswani2017AttentionIA}. 
We also conducted experiments using BERT as the classifier and GPT-2 as the language model, 
but the results are worse than XLNet because BERT and GPT-2 do not share the same vocabulary. 

In this paper, we also conducted experiments with LSTM-based language models. 
However, the learned prior given by XLNet cannot be directly used by LSTM-based language models. 
Since XLNet uses SentencePiece 
\cite{Kudo2018SentencePieceAS} to tokenize the text, each token of LSTM-based language models may be divided into 
multiple sub-tokens. 
To solve this problem, we use the label of the first sub-token as the label for the given token. 
For example, the token ``fine-tuning'' is divided into four sub-tokens 
(``\_fine'', ``-'', ``tun'' and ``ing'') by XLNet. The label of ``\_fine'' 
will be regarded as the label of 
``fine-tuning'' when feeding the outputs of XLNet classifier to LSTM-based language models.

After getting the learned prior table $P$ shown in Figure \ref{example}, 
we set the replacement probability of the ``$<$EOS$>$" token and lexical constraints to zero 
to make sure the given keywords appear in the output. 
In addition, we set the probabilities of the ``$<$BOS$>$" token to zero 
to prohibit any modification at the starting position. 
We computed the sum of these probabilities across the horizontal dimension of the table to compute the probabilities for the replacement and insertion actions, 
from which we sampled an action $a$. Then, we drew a position from the row of the sampled action. 

\subsection{The Candidate Generator and MCMC Sampling}
Suppose the candidate sentence at time step $t$ is $X_t=[x_1,\cdots,x_i,\cdots,x_n]$. 
Assume the proposed action and position is replacement and $i$, respectively. 
We need to compute the probability of replacing the $i$-th token with another token. 
Using Gibbs sampling \cite{gibbs}, we computed the conditional distribution as follows:
\begin{eqnarray}\label{eq:5}
  q(x_i=w_j|x_{-i}) =\frac{p(x_1,\cdots,x_i=w_j,\cdots,x_n)}{ \sum_{w\in V}p(x_1,\cdots,x_i=w,\cdots,x_n)},
\end{eqnarray} 
where $V$ denotes the vocabulary, and $x_{-i}$ denotes tokens of $X_t$ except $x_i$. 
$p(x_1,\cdots,x_i=w_j,\cdots,x_n)$ denotes the sentence probability, computed by a forward 
language model. 

It is non-trivial to compute the conditional distribution with Equation (\ref{eq:5})
since we need to compute probabilities for $|V|$ candidate sentences. 
Previous work resorted to a candidate generator to solve this problem, which takes $x_{-i}$ as inputs and 
outputs the top $K$ tokens with the highest probabilities. Then, we get $K$ candidate sentences
by replacing the $i$-th token with the top $K$ tokens. 
In this paper, we test four different candidate generators. 
First, we trained a forward language model (FLM), a backward language model (BLM), 
and a masked language model (MLM). 
FLM takes the tokens before $x_i$ as inputs ($FLM(x_{<i})$), while BLM takes the tokens after 
$x_i$ as inputs ($BLM(x_{>i})$). To leverage both the past and future contexts, 
MLM takes the tokens before and after $x_i$ as inputs ($MLM(x_{<i}, x_{>i})$). 
Our last method combines FLM and BLM ($FLM(x_{<i})\times BLM(x_{>i})$), 
and also uses the contextual information to predict $x_{i}$. 
Not surprisingly, the last method performs better than using only FLM or BLM as the candidate generator. 
However, we also found that the last method performs slightly better than MLM.  
We speculated that the conditional distribution given by the last method is more consistent with Equation (\ref{eq:5}). 
Therefore, in the following experiments, we use the last method as our candidate generator, 
and we set $K$ to 50.

\begin{algorithm}[h]
  \caption{Constrained Sentence Generation with XLNet}
  \label{alg}
      \begin{algorithmic}[1]
          \STATE Create the synthetic dataset, and train a token-level classifier $C$ on the synthetic training set.
          \STATE Train FLM and BLM on the training set.
          \STATE Set lexical constraints as the initial state $X_0$ of MCMC.
          \FOR {$t \gets 1$ to $T$} 
            \STATE Compute the learned prior $C(X_{t-1})$, and draw an action $A$ and a position $P$ from $C(X_{t-1})$. 
            \STATE Compute the top $K$ candidate tokens with the candidate generator for $(A,P,X_{t-1})$.
            \STATE Create $K$ candidate sentences with the top $K$ tokens, and compute probabilities for them with FLM.
            \STATE Draw a sentence $X^{\prime}$ from the candidate sentences based on the normalized sentence probabilities.
            \STATE Compute the acceptance rate $A(X^{\prime}|X_{t-1})$, and draw a number $\alpha$ from Uniform [0,1]. 
            \STATE \textbf{If} {$\alpha < A(X^{\prime}|X_{t-1})$} \textbf{then} {$X_t$ $\gets$ $X^{\prime}$} \textbf{else} {$X_t$ $\gets$ $X_{t-1}$}.
          \ENDFOR
          \STATE Output the sentence $X^{*}$ with the lowest NLL.
      \end{algorithmic}
\end{algorithm}

Assume the proposed action is insertion. We need to insert a special token, ``$<$MASK$>$", before $x_i$. 
We used the candidate generator to generate 
the top $K$ candidate sentences. 
Then, we used FLM to compute sentence probabilities for them. 
Finally, we sampled a sentence $X^{\prime}$ from them. Similar to the replacement action, 
a sentence with a higher probability is more likely to be selected. 
The difference is that the proposal distribution for insertion is asymmetric. 
To make it meet the \textit{detailed balance condition}, the Metropolis-Hastings (MH) algorithm \cite{metropolis1953equation,hastings1970monte, he2017fastbtm, he2017optimize} 
introduces an acceptance term: 
\begin{eqnarray}\label{eq:6}
  A_{insert}(X^{\prime}|X_t)&=&min(1,A^{\star}_{insert}(X^{\prime}|X_t)), \\
  A^{\star}_{insert}(X^{\prime}|X_t) &\approx& \frac{  q(X_t|X^{\prime})\times p(X^{\prime})}{ q(X^{\prime}|X_t)\times p(X_t)} \\
  &\approx& \frac{ \sum_{X\in S}p(X)}{p(X_t)},
\end{eqnarray}
where $S$ is the set of candidate sentences for the insertion action. 
When computing the acceptance rate, 
we ignored the probabilities for the chosen action and position, 
and we found ignoring these terms improves the experiment results. 

We summarize the process of our proposed approach in Algorithm \ref{alg}. 
We first created synthetic datasets and then fine-tuned XLNet on them to get the classifier. 
Next, we trained FLM and BLM and used them as the candidate generator. 
Finally, we refined the candidate sentence with the classifier and MCMC sampling.

\section{Experiments}
\subsection{Generating Sentences with Lexical Constraints}
\subsubsection{Experiment Setups and Baselines.} 
We selected 6M, 0.3M and 1K sentences from One-Billion-Word corpus as the training, validation, and test sets, respectively. 
We trained forward and backward LSTM-based language models on the training set. 
Similarly, we fine-tuned the pre-trained XLNet (base-cased version) model on the training set 
to get forward and backward XLNet-based language models. 
For each language model, we selected the checkpoint with the lowest validation loss. 
For both LSTM-based and XLNet-based models, the forward and backward language models serve as candidate generators. 
The forward language models are also used to compute sentence probabilities. 
We fine-tuned XLNet (base-cased version) on the synthetic dataset to get our classifier. 
The experiment setups for language models and the classifier are shown in the Appendix.
We constructed four kinds of test sets for constrained sentence generation
by varying the length of lexical constraints $k$ from 1 to
4. For each case, we randomly extracted $1,000$ sets of lexical
constraints from the test sentences.

To compare with previous work, we implemented two variants of the backward and forward language model 
(sep-B/F and asyn-B/F), and GBS. 
After these models are well-trained, we ran beam search decoding (beam width = 10) to generate sentences with the extracted lexical constraints. 
As for CGMH, we used the code and the pre-trained model provided by the author to generate sentences. 

In this paper, we proposed four models. The LSTM-based MCMC model (L-MCMC) uses LSTM-based models as the generator 
and language models. 
L-MCMC w/ deletion is our implementation for CGMH. 
As shown in Table \ref{tab:result2}, sentences generated by our implementation 
have lower NLL values indicating higher quality. 
L-MCMC only uses replacement and insertion actions, which outperforms L-MCMC w/ deletion. 
Therefore, all our proposed models don't use the deletion action. 
When refining the generated sentence, 
L-MCMC randomly chooses a position and an action. 
In comparison, the LSTM-based MCMC w/ classifier (L-MCMC-C) samples a position and an action from the 
learned prior table provided by the classifier. 
Similarly, we also extended MCMC sampling to XLNet-based models and proposed the XLNet-based MCMC (X-MCMC) and 
the XLNet-based MCMC w/ classifier (X-MCMC-C). 
Our models are implemented with HuggingFace \cite{Wolf2019HuggingFacesTS}. 

For a fair comparison, all MCMC-based models were run for 200 steps to generate sentences with the given constraints and 
then output the sentence with the lowest NLL. 
In addition, sep-B/F, asyn-B/F, GBS, L-MCMC, and L-MCMC-C use the same LSTM model structure. 

\begin{table*}[t] 
  \footnotesize
  \centering
    \begin{tabular}{
     m{0.2\textwidth}<{\centering}|
     m{0.04\textwidth}<{\centering}
     m{0.04\textwidth}<{\centering}
     m{0.04\textwidth}<{\centering}
     m{0.04\textwidth}<{\centering}|
     m{0.04\textwidth}<{\centering}
     m{0.04\textwidth}<{\centering}
     m{0.04\textwidth}<{\centering}
     m{0.04\textwidth}<{\centering}|
     m{0.04\textwidth}<{\centering}
     m{0.04\textwidth}<{\centering}
     m{0.04\textwidth}<{\centering}
     m{0.04\textwidth}<{\centering}
     }
    \hline
     \textbf{Metrics} &\multicolumn{4}{c|}{\textbf{NLL (GPT-2) ($\downarrow$)}} &\multicolumn{4}{c|}{\textbf{BLEU (bigram) ($\uparrow$)}}  &\multicolumn{4}{c}{\textbf{Human evaluation ($\uparrow$)}}\\
     \hline
     \textbf{Models} & $k$=1 & $k$=2& $k$=3 & $k$=4  & $k$=1 & $k$=2& $k$=3 & $k$=4 & $k$=1 & $k$=2& $k$=3 & $k$=4\\
     \hline
     sep-B/F  &4.432 &- &- &-  &\textbf{7.1\%} &- &- &-  &0.333 &- &- &- \\
     asyn-B/F &4.304 &- &- &-  &\textbf{7.1\%} &- &- &-   &0.369 &- &- &-\\ 					
     GBS    &3.985	&4.163	&4.178	&4.209 &6.6\%	&\textbf{9.8\%}	&\textbf{12.8\%}	&16.0\% & 0.495&0.384 &0.445 &0.421\\
     CGMH (200)  &5.079	&5.103	&5.227	&5.246 &1.3\%	&3.9\%	&7.3\%	&11.5\%  & 0.345&0.344 &0.400 &0.408\\
     L-MCMC w/ deletion (200) &4.923 &4.775 &4.856 &4.874 &1.0\% &3.9\% &7.7\% &12.5\% &- &- &- &-\\
     \hline
     L-MCMC (200)  &4.549	&4.564	&4.617	&4.672 &2.5\%	&5.9\%	&10.3\%	&14.4\%  &0.433 &0.459 &0.469 &0.429\\
     \textbf{L-MCMC-C} (50)  &4.425	&4.389	&4.491	&4.487 & 1.6\% &4.8\% &8.9\% &14.2\% &- &- &- &-\\
     \textbf{L-MCMC-C} (200)  &3.762	&3.773	&3.819	&3.904 & 3.7\% &7.5\% &12.4\% &\textbf{16.7\%} &0.575 &0.520 &0.573 &0.557\\
     
     \hline
     X-MCMC (200) &4.225	&4.319	&4.427	&4.463 &1.7\%	&4.5\%	&8.3\%	&12.4\%  &0.471 &0.516 &0.487 &0.516\\
     \textbf{X-MCMC-C} (60)   &4.152	&4.259	&4.398	&4.490 &1.9\%	&4.7\%	&8.3\%	&12.1\% &- &- &- &-\\
     \textbf{X-MCMC-C} (200)  &\textbf{3.532}	&\textbf{3.683}	&\textbf{3.779}	&\textbf{3.879} &2.9\%	&6.1\%	&10.0\%	&14.4\% &\textbf{0.681} &\textbf{0.589} &\textbf{0.621} &\textbf{0.655}\\

    \hline
    \hline
     \textbf{Metrics} &\multicolumn{4}{c|}{\textbf{Self-BLEU (4-gram) ($\downarrow$)}} &\multicolumn{4}{c|}{\textbf{Distinct (bigram) ($\uparrow$)}} &\multicolumn{4}{c}{\textbf{Entropy (4-gram) ($\uparrow$)}}\\
     \hline
    Human Reference&8.2\% &9.0\% &9.0\% &9.0\% &81.5\% &80.7\% &80.7\% &80.7\% &9.882 &9.875 &9.875 &9.875\\
    \hline
     sep-B/F  &81.7\% &- &- &-  &19.5\% &- &- &-  &7.664 &- &- &- \\
     asyn-B/F &80.2\% &- &- &-  &20.3\% &- &- &-   &7.924 &- &- &-\\ 					
     GBS &84.2\%	&80.6\%	&73.4\%	&67.9\%   &16.2\%	&20.9\%	&27.0\%	&31.8\% & 6.741&7.555 &8.117 &8.567\\
     CGMH (200)&19.8\%	&14.9\%	&11.7\%	&11.4\%  &65.6\%	&68.0\%	&69.9\%	&71.0\% & 8.605&8.871 &9.124 &9.301\\
    L-MCMC w/ deletion (200)&14.4\% &13.8\% &10.5\% &9.2\% &74.1\%	&74.8\%	&76.2\%	&77.0\%  & 8.309&8.676 &8.961 &9.172\\
    \hline
     L-MCMC (200) &20.1\%	&16.5\%	&13.8\%	&13.3\%  &64.4\%	&67.3\%	&68.6\%	&68.6\% &9.014 &9.260 &9.497 &9.669\\
     \textbf{L-MCMC-C} (50) &20.2\%	&16.3\%	&14.2\%	&12.6\% &68.6\%	&70.8\%	&71.1\%	&73.1\% &8.591 &8.901 &9.144 &9.339\\
     \textbf{L-MCMC-C} (200) &33.0\%	&28.3\%	&24.1\%	&21.9\% &54.1\%	&57.5\%	&60.1\%	&62.3\% &\textbf{9.219} &\textbf{9.415} &\textbf{9.577} &\textbf{9.685}\\
    \hline
     X-MCMC (200) &\textbf{11.3\%}	&\textbf{9.1\%}	&8.3\%	&\textbf{7.1\%}  &\textbf{77.5\%}	&\textbf{78.9\%}	&\textbf{79.4\%}	&79.3\% &8.832 &9.141 &9.391 &9.559\\
     \textbf{X-MCMC-C} (60) &12.6\%	&11.1\%	&\textbf{8.1\%}	&8.5\%  &75.5\%	&77.3\%	&79.3\%	&\textbf{80.0\%} &8.781 &9.018 &9.202 &9.372\\
     \textbf{X-MCMC-C} (200) &19.0\%	&15.1\%	&13.6\%	&12.1\%  &69.6\%	&72.6\%	&74.0\%	&74.2\% &9.097 &9.279 &9.453 &9.577\\
     \hline
  \end{tabular}
  \caption{ Results on One-Billion-Word test sets with different 
  $k$. 
  (Numbers in brackets refer to the number of time steps.)
  }\label{tab:result2}
\end{table*}

\subsubsection{Automatic Evaluation for Quality.} 
To automatically evaluate the quality of generated sentences, 
we followed \cite{wang-cho-2019-bert} by computing negative log-likelihood (NLL) of sentences. 
A lower NLL value means that the generated sentence is more fluent and coherent. 
We used the pre-trained GPT-2 small (117M) to measure NLL values of sentences. 
We also computed BLEU scores \cite{Papineni2002BleuAM} between generated sentences and human references. 
A high BLEU score indicates a model can generate sentences similar to human references. 
NLL and BLEU results are shown in Table \ref{tab:result2}. 
Our proposed models (L-MCMC-C and X-MCMC-C) outperform baselines in NNL. 

It is worth mentioning that sep-B/F, asyn-B/F, and GBS achieve relatively low NLL 
and high BLEU scores. 
Does it mean these models can generate well-formed sentences? 
Holtzman et al. \shortcite{Holtzman2020TheCC} found that beam search decoding tends to 
lead to degeneration. Language models are expected to assign low NLL scores to high-quality sentences, 
yet they also give low NLL scores to repetitive and generic sentences. 
We found that beam search-based models (sep-B/F, asyn-B/F, and GBS) may easily 
fall into repetitive loops, which is consistent with what is observed in previous work \cite{Holtzman2020TheCC}. 
We showed the percentage of sentences containing n-gram repetitions in Table \ref{tab:repetition} in the Appendix.
In addition, we found that some generic sub-sequences 
(`I am going to', `he said', `referring to', `adding that', etc.) frequently appear in the generated sentences. 
Both repetitive n-grams and generic sub-sequences will help these models to achieve low NLL and high BLEU scores. 
 Therefore, it is insufficient to assess the generated sentences with NLL and BLEU. 
 To complement them, we also automatically measured the diversity of each model’s generations.

\begin{figure*}
  \centering
    \subfigure[Acceptance rates vs. $k$.]{
      \centering
      \begin{minipage}{5.1cm}
        \includegraphics[width=1\textwidth]{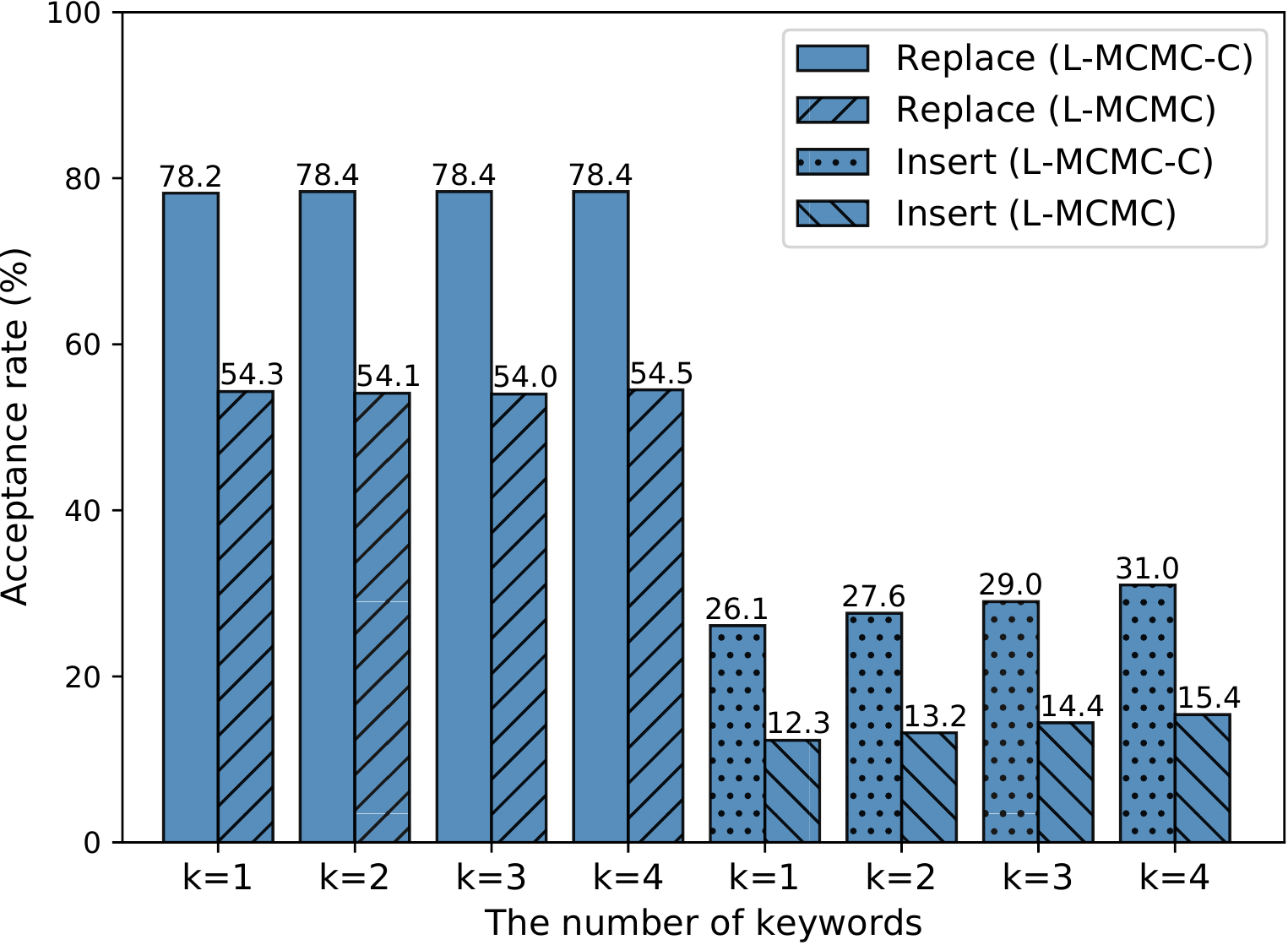} 
      \end{minipage}
    }
      \subfigure[NLL vs. Self-BLEU.]{
        \centering
        \begin{minipage}{5.7cm}
        \includegraphics[width=1\textwidth]{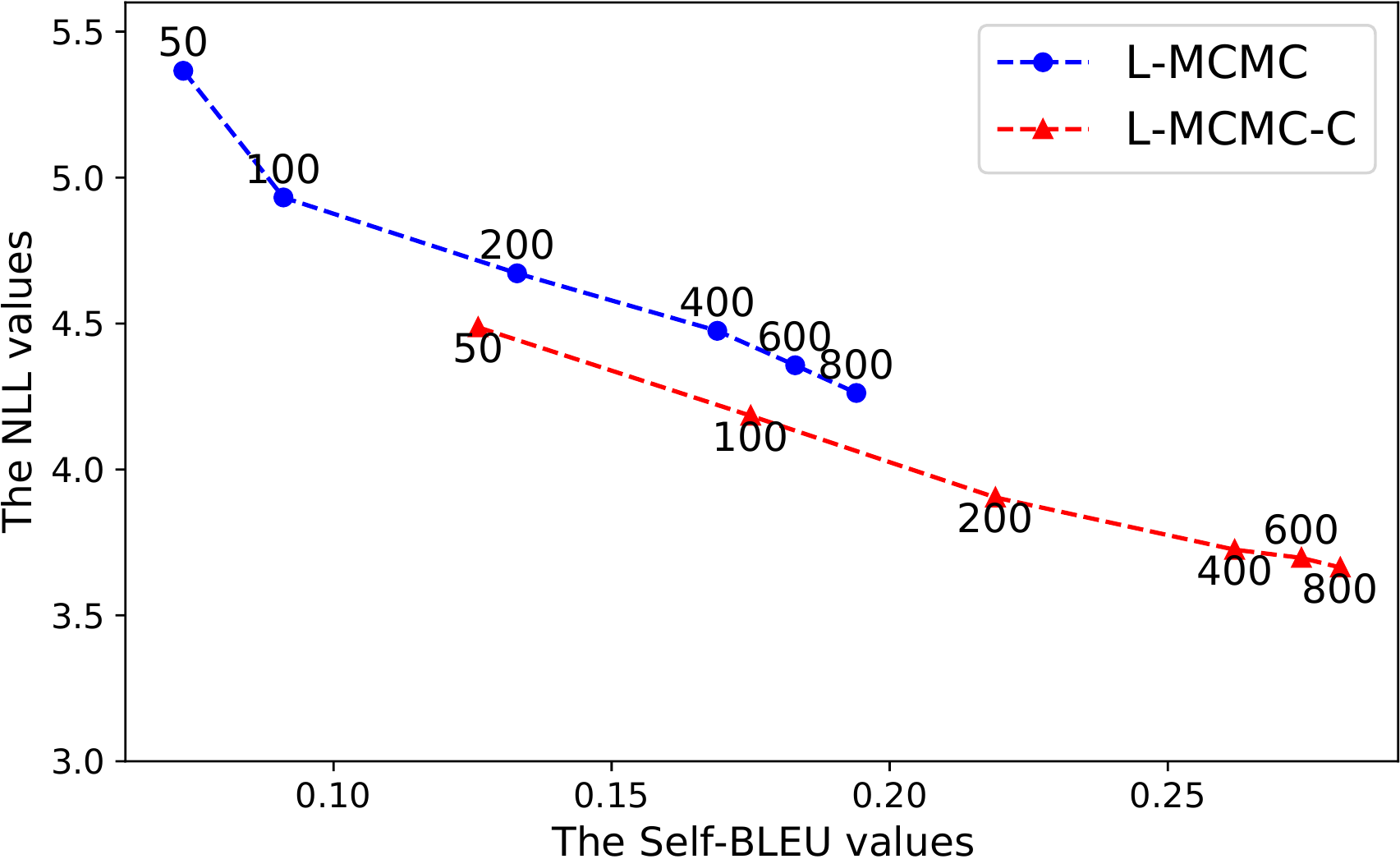} 
        \end{minipage}
      }
      \subfigure[NLL and Self-BLEU vs. time steps.]{
        \centering
        \begin{minipage}{6cm}
        \includegraphics[width=1\textwidth]{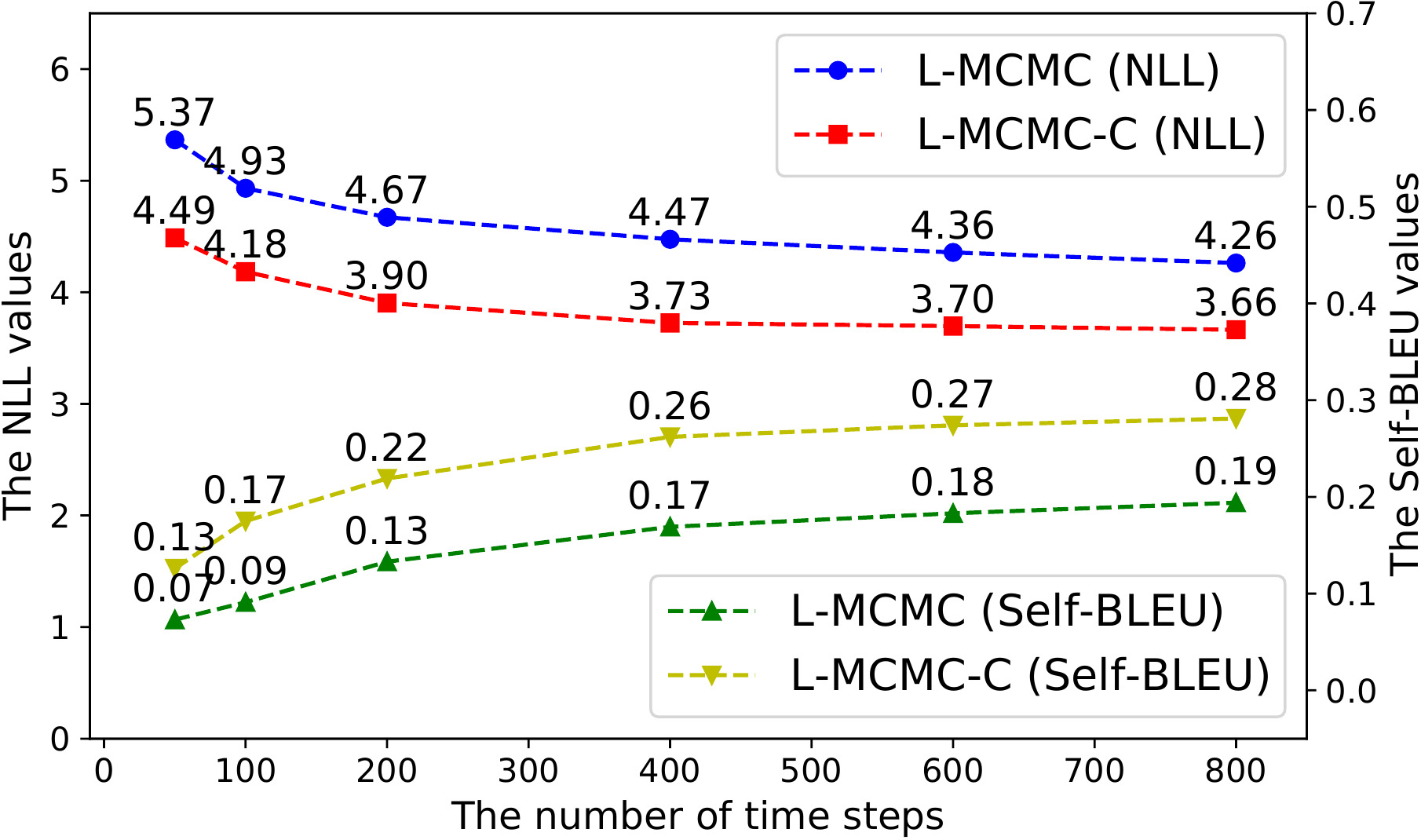} 
        \end{minipage}
      }
      \caption{ In subfigure (a), $k$ is the number of keywords. 
      In subfigure (b), numbers near lines denote the number of time steps.
      }
      \label{steps}
\end{figure*}

\subsubsection{Automatic Evaluation for Diversity.} 
We used Self-BLEU \cite{Zhu2018TexygenAB} to measure the diversity of each model’s generations.
Self-BLEU evaluates how one sentence resembles the other generated ones. For each sentence, we treated it as 
 a hypothesis and the others as references. 
We also used distinct n-gram \cite{Li2016ADO} 
and entropy \cite{Zhang2018GeneratingIA} to measure diversity.
Distinct n-gram reflects the percentage of unique n-grams. 
Entropy further considers the distribution of n-grams. 
Lower Self-BLEU or higher distinct n-gram and entropy values indicate higher diversity.
From Table \ref{tab:result2}, we can see that sep-B/F, asyn-B/F and GBS 
have higher Self-BLEU, lower distinct bigram and lower entropy scores, 
indicating they have low sample diversity, which is consistent with our analysis above.
By comparison, MCMC-based models have high sample diversity. 

When we ran L-MCMC-C for 50 steps, it achieved similar performance in sentence quality, and diversity to L-MCMC ran for 200 steps.  
Similarly, X-MCMC-C with 60 refinement steps is on par with X-MCMC with 200 refinement steps. 
We showed NLL and Self-BLEU achieved by L-MCMC and L-MCMC-C with different refinement steps when $k=4$ in Figure \ref{steps}(b). 
We can see that the proposed model (L-MCMC-C) requires much fewer refinement steps to achieve similar 
sentence quality and diversity compared to L-MCMC. 
The improvements achieved by our proposed models are mainly due to the classifier's guidance, 
which reduces the number of invalid refinements. 
To verify this point, 
we showed the acceptance rates of LSTM-based models in Figure \ref{steps}(a)
and showed the acceptance rates of XLNet-based models in the Appendix. 
We can see that applying the classifier significantly improves the acceptance rates 
for replacement and insertion actions. In addition, 
adding the classifier brings a small amount of overhead. 
For example, the running time of the classifier only accounts for about 1/7 of X-MCMC-C. 

When we ran L-MCMC-C and X-MCMC-C for 200 steps, NLL values declined dramatically. 
Meanwhile, Self-BLEU increased, and distinct bigram decreased. 
Increasing the number of time steps will improve sentence quality but degrade sentence diversity. 
It seems to indicate that sentence quality and diversity contradict each other. 
To further analyze this phenomenon, we showed the relationship between NLL and Self-BLEU when $k=4$ in Figure \ref{steps}(c). 
We can see that NLL values decrease with the increase of time steps, 
while Self-BLEU values increase for both models. 
Therefore, it is nontrivial to improve both sentence quality and diversity. 
Generic sentences with high probabilities favor sentence quality in terms of NLL but disfavor sentence diversity. 
Another advantage of our method is that we can make a trade-off between quality and diversity by controlling 
the number of time steps. 

When comparing L-MCMC with X-MCMC (or L-MCMC-C with X-MCMC-C), 
we found the latter outperforms the former in both sentence quality (NLL) and diversity (Self-BLEU and distinct bigram), 
mainly because the latter uses the pre-trained model as language models. 

\subsubsection{Human Evaluation.} 
We also conducted a human evaluation to further compare these models. 
We selected 100 sentences for each model and invited three volunteers to assess the sentences in terms of fluency. 
Concretely, we first chose a group of sentences generated by different models and shuffled them to avoid bias. 
Then, the annotators ranked these sentences and assigned scores (0 to 1) to them, where 1 indicates the
best quality. 
We showed the results of human evaluation in Table \ref{tab:result2}. 
Our proposed models (X-MCMC-C and L-MCMC-C) also outperform baselines in human evaluation. 
We also performed paired t-test comparisons between the proposed model (X-MCMC-C)
 and baselines, and $p$-values are less than $0.01$, 
 indicating the differences between the proposed model and baselines are statistically significant.

\subsubsection{Ablation Study.}

We performed an ablation study to demonstrate the importance of each design. 
We mainly focused on two aspects: 
the effectiveness of the learned prior and the importance of the synthetic dataset. 
In Table \ref{tab:result3}, we compared L-MCMC-C (row 1) with five variants (rows 2-6). 
For a fair comparison, all models use LSTM-based language models and are run for 200 steps. 

Removing the learned prior for actions (row 2)
(actions are randomly sampled, but positions are still sampled from the learned prior) 
results in a slight increase in NLL, 
while removing the learned prior for positions (row 3) causes a sharp decline in performance,
 which is nearly as poor as L-MCMC (row 7). 
 Therefore, the learned prior for positions is much more important than that of actions. 
Further, we used one-third of the synthetic dataset to train the classifier (row 4). 
The results of row 4 are slightly worse than those of row 1. Based on row 4, if we remove 
the random method and only use the masked LM (XLNet) to create the synthetic dataset (row 5), 
NLL values will increase marginally. By comparison, removing the masked LM (row 6) has a 
detrimental effect on performance. The random and masked LM methods are complementary, 
and both are indispensable for training the classifier. 

To summarize, both the learned prior and the synthetic dataset play an essential role in our proposed model.

\begin{table}
    \footnotesize
  \centering
    \begin{tabular}{
    m{0.01\textwidth}<{\centering}
     m{0.16\textwidth}<{\raggedright}|
     m{0.04\textwidth}<{\centering}
     m{0.04\textwidth}<{\centering}
     m{0.04\textwidth}<{\centering}
     m{0.04\textwidth}<{\centering}
     }

    \hline
    ~&\textbf{Metrics} & \multicolumn{4}{c}{\textbf{NLL (GPT-2) ($\downarrow$)}} \\
    \hline
    \# & \textbf{Models variants} & $k$=1 & $k$=2& $k$=3 & $k$=4 \\
     \hline
     
    1& \textbf{L-MCMC-C} &\textbf{3.762}	&\textbf{3.773}	&\textbf{3.819}	&\textbf{3.904}\\
    \cline{2-6}
    2&\quad-- prior for actions &3.850	&3.861	&3.893	&3.932\\
    3&\quad -- prior for positions &4.498 &4.513 &4.605 &4.646\\
    4&\quad -- size of dataset &3.921	&3.950	&4.015	&4.030\\
    5&\qquad-- random &3.954	&3.982	&4.024	&4.066\\
    6&\qquad-- masked LM &4.063	&4.108	&4.167	&4.189\\
    \cline{2-6}
    7& L-MCMC &4.549	&4.564	&4.617	&4.672\\
    \hline
  \end{tabular}
  \caption{Ablation study on our model. 
  }\label{tab:result3}
\end{table}

\begin{table}[t]
    \scriptsize
  \centering
  \begin{tabular}{
    m{0.08\textwidth}<{\centering}|
    m{0.34\textwidth}<{}
    }
    \hline
    \textbf{Constraints} & \textbf{person}, \textbf{home}, \textbf{problems}, \textbf{depression}\\
    \hline
    GBS & A \textbf{person} familiar with the matter said : `` The \textbf{problems} of \textbf{depression} are \textbf{home} to the people of the United States .\\
    \hline
    CGMH& The \textbf{person} 's head of \textbf{home} health \textbf{problems} have sparked \textbf{depression} .\\
    \hline
    \textbf{L-MCMC-C} & The average \textbf{person} who 's lost money since he was a child at \textbf{home} has no \textbf{problems} with \textbf{depression} .\\
    \hline
    \textbf{X-MCMC-C}& One \textbf{person} was sent \textbf{home} with mental health \textbf{problems} and severe \textbf{depression} .\\
    \hline
  \end{tabular}
  \caption{Sentences generated with lexical constraints.}
  \label{tab:case1}
\end{table}

\begin{table*}[!t] 
  \footnotesize
  \centering
    \begin{tabular}{
     m{0.19\textwidth}<{\centering}|
     m{0.039\textwidth}<{\centering}
     m{0.039\textwidth}<{\centering}
     m{0.039\textwidth}<{\centering}
     m{0.039\textwidth}<{\centering}
     m{0.039\textwidth}<{\centering}
     m{0.039\textwidth}<{\centering}|
     m{0.039\textwidth}<{\centering}
     m{0.039\textwidth}<{\centering}
     m{0.039\textwidth}<{\centering}
     m{0.039\textwidth}<{\centering}
     m{0.039\textwidth}<{\centering}
     m{0.039\textwidth}<{\centering}
     }

    \hline
    \textbf{Metrics} &\multicolumn{6}{c|}{\textbf{NLL (GPT-2) ($\downarrow$)}} &\multicolumn{6}{c}{\textbf{BLEU (4-gram) ($\uparrow$)}} \\
    \hline
    \multirowcell{2}{\textbf{Models}}&
    \multicolumn{3}{c}{Middle}& 
    \multicolumn{3}{c|}{Random}&
    \multicolumn{3}{c}{Middle}&
    \multicolumn{3}{c}{Random}\\
    \cline{2-13}
     ~ & 25\% & 50\% & 75\% & 25\% & 50\% & 75\% & 25\% & 50\% & 75\% & 25\% & 50\% & 75\%\\
     \hline
    Human Reference&4.022&4.022&4.022&4.022&4.022&4.022 &100\%&100\%&100\%&100\%&100\%&100\% \\
    Template &-&-&-&-&-&- &65.1\% &36.5\% &9.0\% &49.5\% &16.1\% &2.6\% \\
    \hline
    FLM &4.417&4.479 &4.223 &4.434&4.727&4.543 &68.4\% &39.5\% &12.2\% &66.4\% &31.9\% &10.5\%\\
    BLM &4.453 &4.460 &4.130 &4.545  &4.860 &4.620 &68.3\% &39.6\% &12.8\% &65.9\% &31.2\% &9.9\%\\
    F+B LM & 4.288 &4.355 &4.130 &4.309  &4.605 &4.457 &\textbf{69.1}\%&40.0\% &12.5\% &68.3\% &33.4\% &\textbf{10.9\%}\\
    Bayesian MCMC &4.188 &4.268 &4.242 &4.164 &4.310 &4.349 &68.7\% &39.6\% &12.3\% &69.7\% &35.4\% &10.5\%\\
    TIGS (strategy 1) & 4.985 &5.515 &5.865 &5.183 &5.911 &6.268 &65.8\% &37.2\% &9.9\% &56.1\% &21.8\% &4.3\%\\
    TIGS (strategy 2) & 4.416 &4.341 &4.042 &4.638  &5.018 &4.852 &67.9\% &39.7\% &12.3\% &62.9\% &28.5\% &7.9\%\\
    \hline
    L-MCMC &4.216 &4.313 &4.303 &4.178 &4.367 &4.436 &68.5\% &39.7\% &12.5\% &69.0\% &34.4\% &10.1\%\\
    \textbf{L-MCMC-C} &\textbf{4.146} &\textbf{4.053} &\textbf{3.796} &\textbf{4.135} &\textbf{4.199} &\textbf{4.106} &69.0\% &\textbf{40.3\%} &\textbf{12.9\%} &\textbf{70.0\%} &\textbf{36.1\%} &10.5\%\\
    \textbf{L-MCMC-C (w/o BP)}&-&-&-&-&-&- &70.9\%&43.1\%&14.6\%&70.9\%&37.6\%&11.8\%\\
    \hline
  \end{tabular}
  \caption{NLL and BLEU results for different mask strategies and rates. (``BP" refers to the brevity penalty.)
  }\label{tab:result4}
\end{table*}

\subsubsection{Samples and Analysis.}
We showed some sentences generated by our proposed models and baselines in Table \ref{tab:case1}.
Our proposed models can generate high-quality, lexically constrained sentences. 
As for GBS, the lexical constraint ``home'' conflicts with the previous tokens
 since GBS is not aware of the future lexical constraints when generating previous tokens. 
 Therefore, forcing to incorporate the lexical constraint ``home'' degrades the quality of the generated sentence. 
The sentence generated with CGMH lacks coherence and fluency. Since CGMH refines the generated sentence randomly, it still needs more refinements.
More generated sentences are shown in Table \ref{tab:case2} in the Appendix. 
\subsection{Text Infilling}
\subsubsection{Experiment Setups and Baselines.}
We used $1,000$ sentences from the test set of One-Billion-Word corpus to create test sets for text infilling. 
Following \cite{Liu2019TIGSAI}, we resorted to two mask strategies (random and middle) and three mask ratios 
($r$ = 25\%, 50\%, or 75\%) to construct test sets for text infilling. 
The random mask strategy randomly removes $r$ 
of tokens from the original sentence. 
The middle mask strategy removes $r$ of tokens from the middle of the original sentence. 
Therefore, we have six types of test sets. 

To compare our proposed model with previous work, we implemented several strong baselines. 
The forward language model (FLM) and the backward language model (BLM) 
fills the blanks from left to right, and from right to left with beam search (beam width = 10), respectively. 
F+B LM \cite{Wang2016ImageCW} fills the blanks by FLM and BLM and then selects the output with the lowest NLL.
Bayesian MCMC \cite{berglund2015bidirectional} initializes the blanks with random values
and uses the replacement action to refine the filled values, 
but it can only generate sentences with a fixed length. 
We also implemented TIGS \cite{Liu2019TIGSAI}, which iteratively refines the filled tokens with gradient descent.
Before refining with TIGS, we resorted to two strategies to 
initialize the blanks with random values or values predicted by FLM with greedy search. 
All models use LSTM-based language models with the same structure 
(the setups for LSTM-based language models have been introduced in the first task). 
For a fair comparison, we ran 20 iterations for Bayesian MCMC, TIGS, L-MCMC, and L-MCMC-C.
\subsubsection{Automatic Evaluation.}
Following previous work \cite{Liu2019TIGSAI}, we also resorted to NLL and BLEU to automatically 
evaluate the infilled sentences. Similar to our first task, NLL is measured by GPT-2, and 
BLEU is computed between the infilled sentences and human references. 
BLEU measures how similar the infilled sentence is to the ground truth, 
while NLL assesses the fluency and coherence of the infilled sentences. 
We showed the results of NLL and the corpus-level BLEU \cite{Papineni2002BleuAM} in Table \ref{tab:result4}.
Our proposed model achieves the lowest NLL scores in all cases, 
which means the infilled sentences of our proposed model are more fluent than those of previous methods. 
One advantage of our proposed model is that it does not need to know the number of blanks when infilling. 
In contrast, other baselines need to know the number of blanks before infilling. 
Therefore, our proposed models (L-MCMC and L-MCMC-C) may generate sentences 
with different lengths from the ground truths. 
Even though the BLEU algorithm with the brevity penalty will penalize our model 
when it generates shorter sentences, our model still outperforms baselines in most cases in terms of BLEU. 
Compared with TIGS, the proposed model does not need any initialization before infilling. 
In addition, TIGS is sensitive to initialization strategies. TIGS performs much better 
when initialized with the results of FLM, which may be unfair to other models. 
Compared with L-MCMC, 
L-MCMC-C significantly improves the performance in NLL and BLEU, mainly benefiting 
from the learned prior given by the classifier. 
We showed some infilled sentences in Table \ref{tab:case3} in the Appendix. 

\section{Related Work}
\subsection{Pre-trained Language Models} 
Recently, many downstream NLP tasks are driven by large-scale pre-trained language models such as 
GPT \cite{Radford2018ImprovingLU}, GPT-2 \cite{Radford2019LanguageMA}, 
BERT \cite{Devlin2019BERTPO}, ROBERTA \cite{Liu2019RoBERTaAR}, ELECTRA \cite{Clark2020ELECTRAPT}, 
SpanBERT \cite{Joshi2019SpanBERTIP}, XLNet \cite{Yang2019XLNetGA}, 
MASS \cite{Song2019MASS} and BART \cite{Lewis2020BARTDS}. 
However, these models cannot be directly applied to lexically constrained sentence generation. 
\subsection{Generating Sentences with Lexical Constraints} 
B/F LMs \cite{mou2015backward,Liu2019BFGANBA} 
are limited to generating text with one lexical constraint. 
GBS (Hokamp and Liu, 2017) can generate sentences with multiple
lexical constraints but degrades the generation quality
and diversity. 
CGMH \cite{miao2019cgmh} revises candidate sentences 
randomly, causing many invalid operations.
To solve this problem, we used an XLNet-based token-level classifier to guide 
MCMC-based models to refine the candidate sentence.

\section{Conclusion}
In this paper, we aim to reduce the redundant refinements conducted by previous MCMC-based models.
To achieve this, we used a token-level classifier to instruct MCMC-based models where and how to refine 
the candidate sentence. 
Compared with previous MCMC-based approaches, our proposed model can iteratively 
refine the candidate sentence with the learned prior given by the pre-trained classifier. 
Experiment results show that our proposed model can generate fluent and diverse sentences 
for constrained sentence generation, outperforming all baselines.

\section*{Acknowledgements}

We would like to thank the anonymous reviewers for their
constructive and informative feedback. 

\bibliography{aaai}

\clearpage
\appendix{\textbf{Appendix}}

\section{Language Models}\label{appendix_a}

For LSTM-based language models, both the forward and backward language models have two LSTM layers. 
The embedding size and hidden size are $256$.
In addition, we selected 50,000 most frequent tokens of the training set as 
the vocabulary for LSTM-based models. 
 During the training process, we set the dropout to $0.2$ and the learning rate to $1e-4$. 

For XLNet-based language models, we used the pre-trained XLNet (base-cased version) model, 
which has 12 self-attention layers with 110M parameters.
The vocabulary size for XLNet is 32,000. We fine-tuned XLNet on the training set with learning rate $lr=1e-5$. 
We trained all language models on the training set until no improvement on the validation set. 
We selected the best checkpoint with the lowest validation loss. 
 NLL results of the trained language models on the One-Billion-Word validation set are shown in Table \ref{tab:appendix_nll}.

\begin{table}[h] 
  \centering
    \begin{tabular}{
    m{0.15\textwidth}<{\centering}|
     m{0.1\textwidth}<{\centering}
     m{0.1\textwidth}<{\centering}
     }
    \toprule
     \textbf{Models} & \textbf{Forward}& \textbf{Backward}\\
     \midrule
     LSTM-based & 4.272 & 4.448 \\
     XLNet-based &3.076& 3.071  \\
     GPT-2 (small) &3.980 & - \\
    \bottomrule
  \end{tabular}
  \caption{NLL of different language models on the One-Billion-Word validation set. 
   GPT-2 is used to evaluate the quality of the generated sentences. 
  GPT-2 is the pre-trained model without any fine-tuning.
  }\label{tab:appendix_nll} 
\end{table}

\section{Classifier}\label{appendix_b}
We created the synthetic dataset for the copy, replacement, insertion, and deletion actions. 
Similar to the replacement action, we also resorted to two approaches to create synthetic data for the deletion action. 
To create synthetic data for the deletion action, we need to insert some tokens in selected sentences, 
where the inserted tokens are randomly sampled from the vocabulary 
or predicted by the masked language model, i.e., XLNet. 
Both the training and validation sets are created with masked LM and random methods. 
We created 36M sentences as the training set 
(30M sentences are created with the masked LM method, and 6M sentences are created with the random method) and 
1.8M sentences as the validation set 
(1.5M sentences are created with the masked LM method, and 0.3M sentences are created with the random method). 
We fine-tuned the pre-trained XLNet (base-cased version) model on the synthetic training set with learning rate $lr=1e-5$ for two epochs. 
We used precision, recall, and F1 score to evaluate the fine-tuned classifier on the validation set. 
During training, we chose the checkpoint with the best performance on the validation set on the macro-average F1 score.
From Table \ref{tab:appendix_classifier}, we can see that the fine-tuned classifier 
achieves high performance on the random validation set and performs slightly worse on the masked LM validation set, 
mainly because it is more challenging to infer labels for the masked LM validation set. 

In our experiments, 
we only leveraged the fine-tuned classifier to guide MCMC sampling to insert or replace tokens 
without using the deletion action since the deletion action makes the model hesitate to move forward. 
For example, the model may insert a token and then delete it later, 
because at the beginning, both insertion and deletion actions have high probabilities. 
We empirically found that the incomplete sentence can be refined by iteratively inserting tokens.  
Therefore, only using the replacement and insertion actions can generate plausible sentences. 

\begin{figure*}[h]
  \centering
  \subfigure[Example 1]{
  \includegraphics[width=0.45\textwidth]{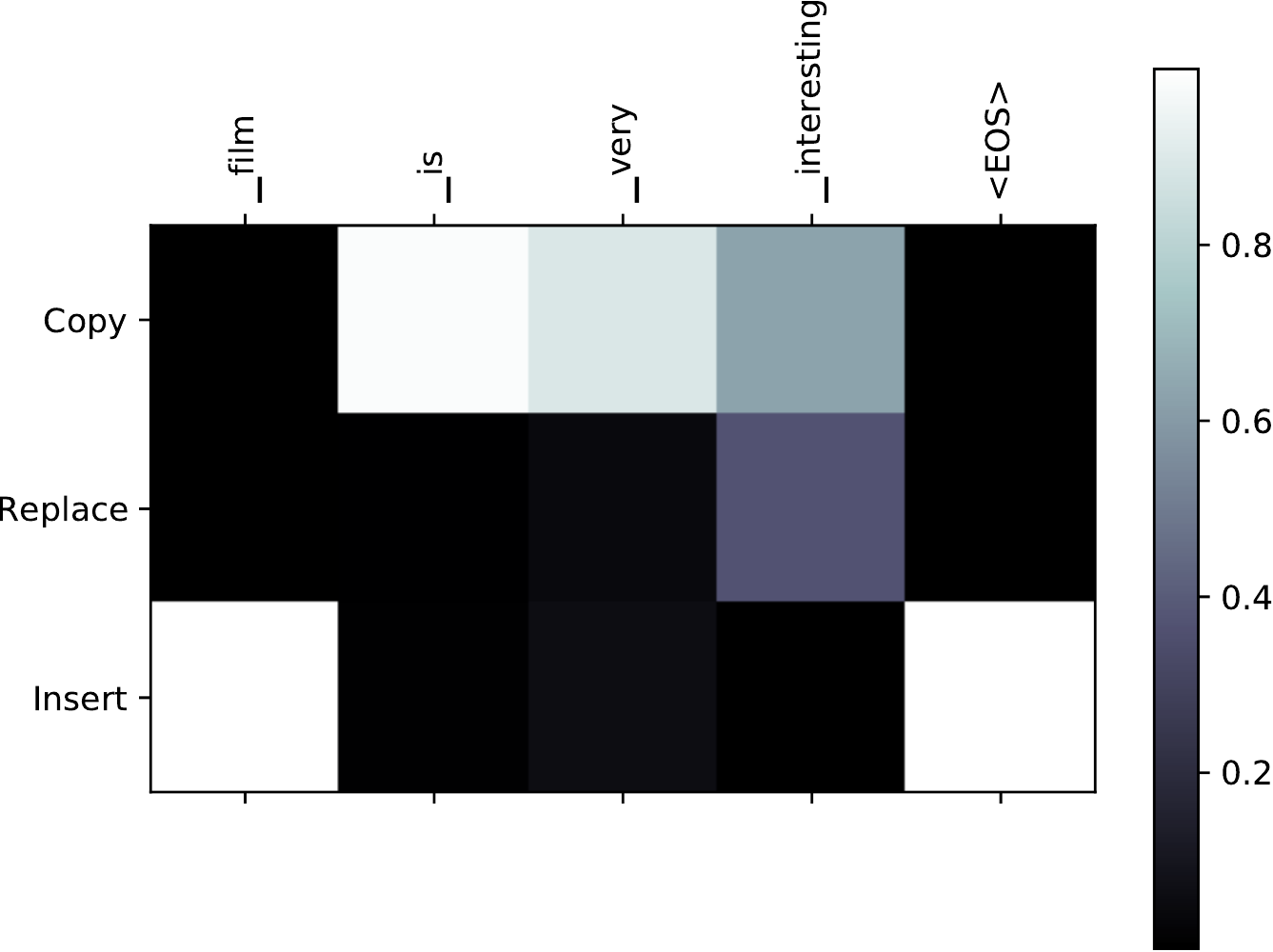} 
  }
  \subfigure[Example 2]{
  \includegraphics[width=0.45\textwidth]{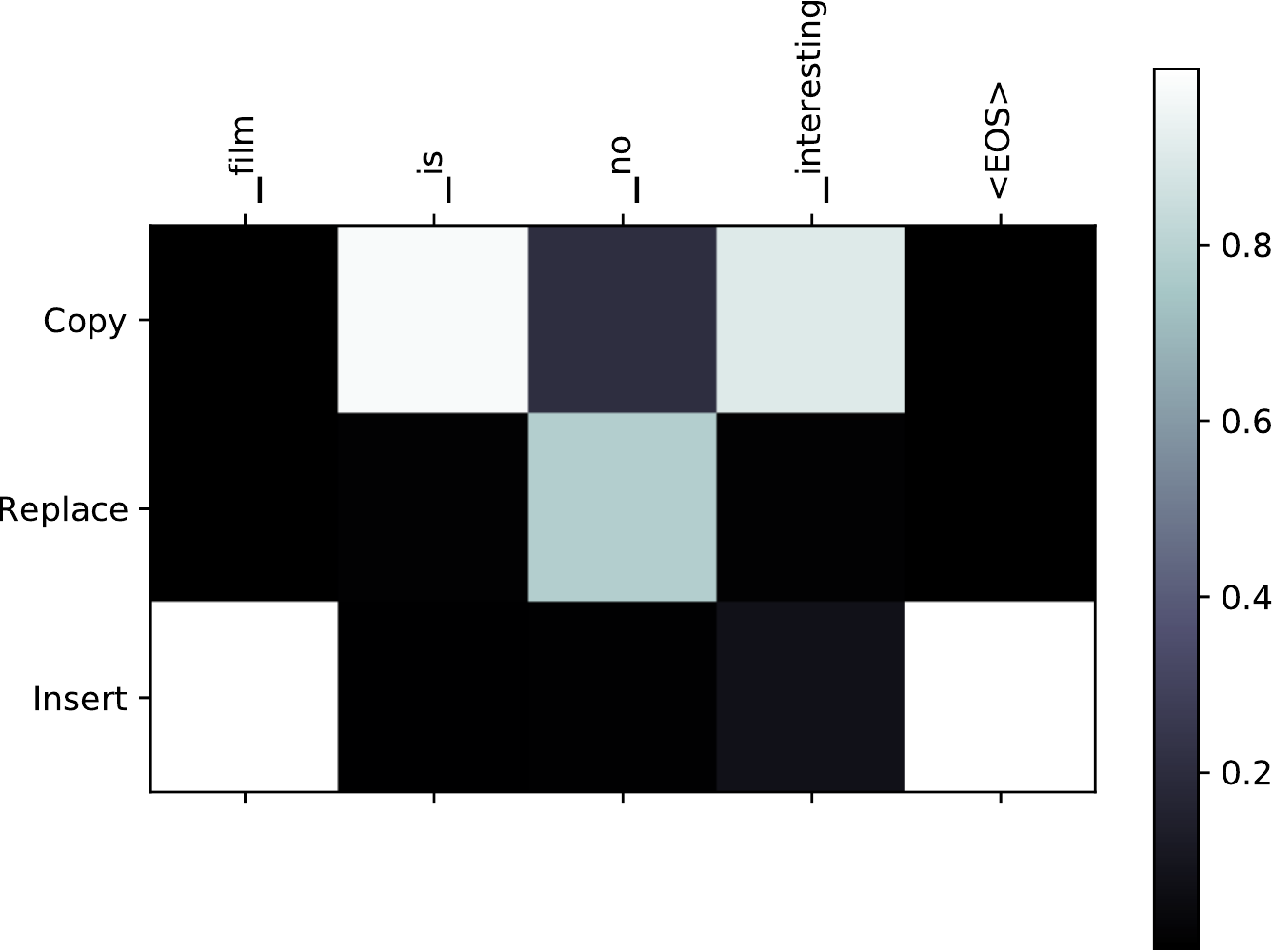} 
  }
  \subfigure[Example 3]{
  \includegraphics[width=0.45\textwidth]{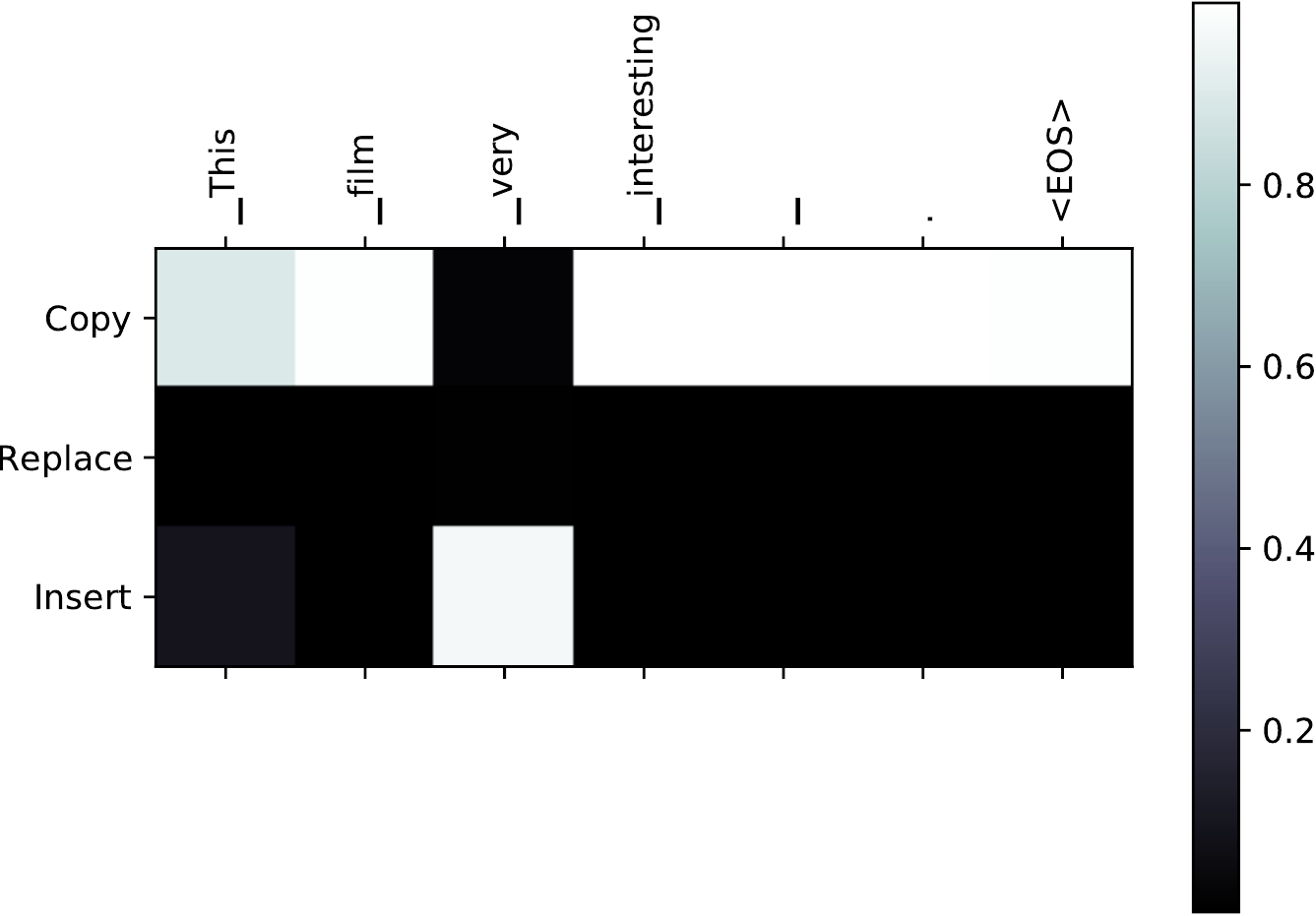} 
  }
  \subfigure[Example 4]{
  \includegraphics[width=0.45\textwidth]{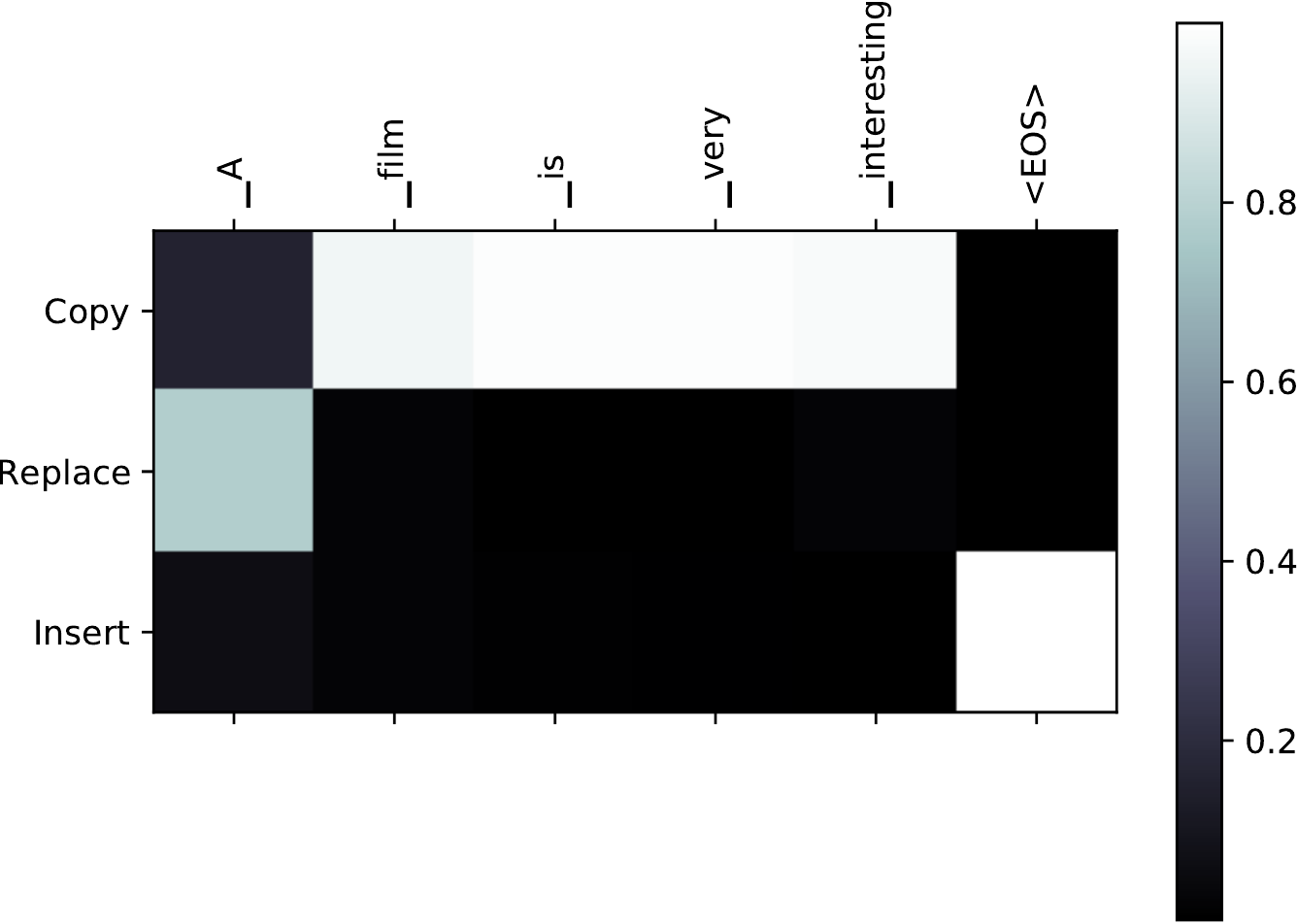} 
  }
  \caption{The learned prior of four different candidate sentences given by the classifier. The x-axis and y-axis of each plot
  correspond to the tokens in the candidate sentence and the actions.}\label{heatmaps}
\end{figure*}

To better understand the classifier's function, we showed the learned prior of four candidate 
sentences in Figure \ref{heatmaps}. In the subfigure (a),  we can see the classifier tells us 
we should insert some tokens before ‘film’ and ‘$<$EOS$>$’. The other tokens should maintain unchanged. 
This prediction is consistent with our intuition. 

\section{N-gram Repetition}
We measured whether a sentence contains n-gram repetitions based on three rules. 
Firstly, if a unigram appears more than three times in a sentence, we regarded the sentence contains a repetition. 
Similarly, if a bigram appears more two times or 
a trigram appears more than one time in a sentence, 
the sentence is also considered as containing a repetition. 
We showed the percentage of sentences containing n-gram repetitions in Table \ref{tab:repetition}. 
All MCMC-based models (CGMH, L-MCMC, L-MCMC-C, X-MCMC, and X-MCMC-C) are run for 200 steps. 
We can see that sentences generated by beam search-based models (sep-B/F, asyn-B/F and GBS) tend to 
get stuck in repetitions. 
\begin{table}[t] 
  \centering
    \begin{tabular}{
     m{0.15\textwidth}<{\centering}|
     m{0.04\textwidth}<{\centering}
     m{0.04\textwidth}<{\centering}
     m{0.04\textwidth}<{\centering}
     m{0.04\textwidth}<{\centering}
     }
    \toprule
     \textbf{Metrics}  &\multicolumn{4}{c}{\textbf{Repetition ($\downarrow$)}}\\
    \midrule
     \textbf{Models} & $k$=1 & $k$=2& $k$=3 & $k$=4  \\
     \midrule
    Human Reference&3.4\% &4.3\% &4.3\% &4.3\%  \\
    \midrule
     sep-B/F  &32.4\% &- &- &-  \\
     asyn-B/F &39.6\% &- &- &-  \\ 					
     GBS &21.2\%	&24.8\%	&21.6\%	&26.1\% \\
     CGMH  &0.5\%	&0.2\%	&0.2\%	&0.2\% \\
     \midrule
     L-MCMC  &0.1\%	&0	&0	&0.2\%\\
     \textbf{L-MCMC-C}  &0.3\%	&0.7\%	&0.4\%	&0.2\%\\
    \midrule
     X-MCMC  &0	&0	&0	&0\\
     \textbf{X-MCMC-C}   &0.1\%&0.2\%	&0.2\%	&0 \\
    \bottomrule
  \end{tabular}
  \caption{The percentage of sentences containing n-gram repetitions. 
  ($k$ denotes the number of lexical constraints.) 
  }\label{tab:repetition}
\end{table}

\begin{table*}[t] 
  \centering
    \begin{tabular}{
    m{0.18\textwidth}<{\centering}|
     m{0.06\textwidth}<{\centering}
     m{0.06\textwidth}<{\centering}
     m{0.06\textwidth}<{\centering}|
     m{0.06\textwidth}<{\centering}
     m{0.06\textwidth}<{\centering}
     m{0.06\textwidth}<{\centering}|
     m{0.06\textwidth}<{\centering}
     m{0.06\textwidth}<{\centering}
     m{0.06\textwidth}<{\centering}
     }
    \toprule
    ~ &\multicolumn{3}{c|}{Masked LM validation set}&\multicolumn{3}{c|}{Random validation set}&\multicolumn{3}{c}{Whole validation set}\\
    \midrule
     Labels & P ($\uparrow$) &R ($\uparrow$)& F1 ($\uparrow$) & P ($\uparrow$)&R ($\uparrow$)& F1 ($\uparrow$)& P ($\uparrow$)&R ($\uparrow$)& F1 ($\uparrow$)\\
     \midrule
     Copy &0.978 & 0.993 & 0.986 &0.989	&0.996	&0.993 &0.980	&0.994	&0.987\\
     Replacement &0.772	&0.462	&0.578 &0.872	&0.801	&0.835&0.795	&0.519	&0.628\\
     Insertion &0.951	&0.893	&0.921 &0.947	&0.894	&0.920 &0.951	&0.893	&0.921\\
     Deletion &0.809	&0.684	&0.741 &0.875	&0.860	&0.868 &0.821	&0.714	&0.764\\
     Macro-average &0.878	&0.758	&0.807 &0.921	&0.888	&0.904 &0.887	&0.780	&0.825\\      
    \bottomrule
  \end{tabular}
  \caption{Results of the classifier on the synthetic validation sets. ``P" and ``R" denote precision and recall. 
  The random validation set is created by replacing some tokens with random tokens. 
  The masked LM validation set is created by the masked LM method. 
  The whole validation set combines the random and masked LM validation sets.
  }\label{tab:appendix_classifier}
\end{table*}

\section{Acceptance Rates of XLNet-based Models}
\begin{figure}
  \centering
    \centering
    \includegraphics[width=0.42\textwidth]{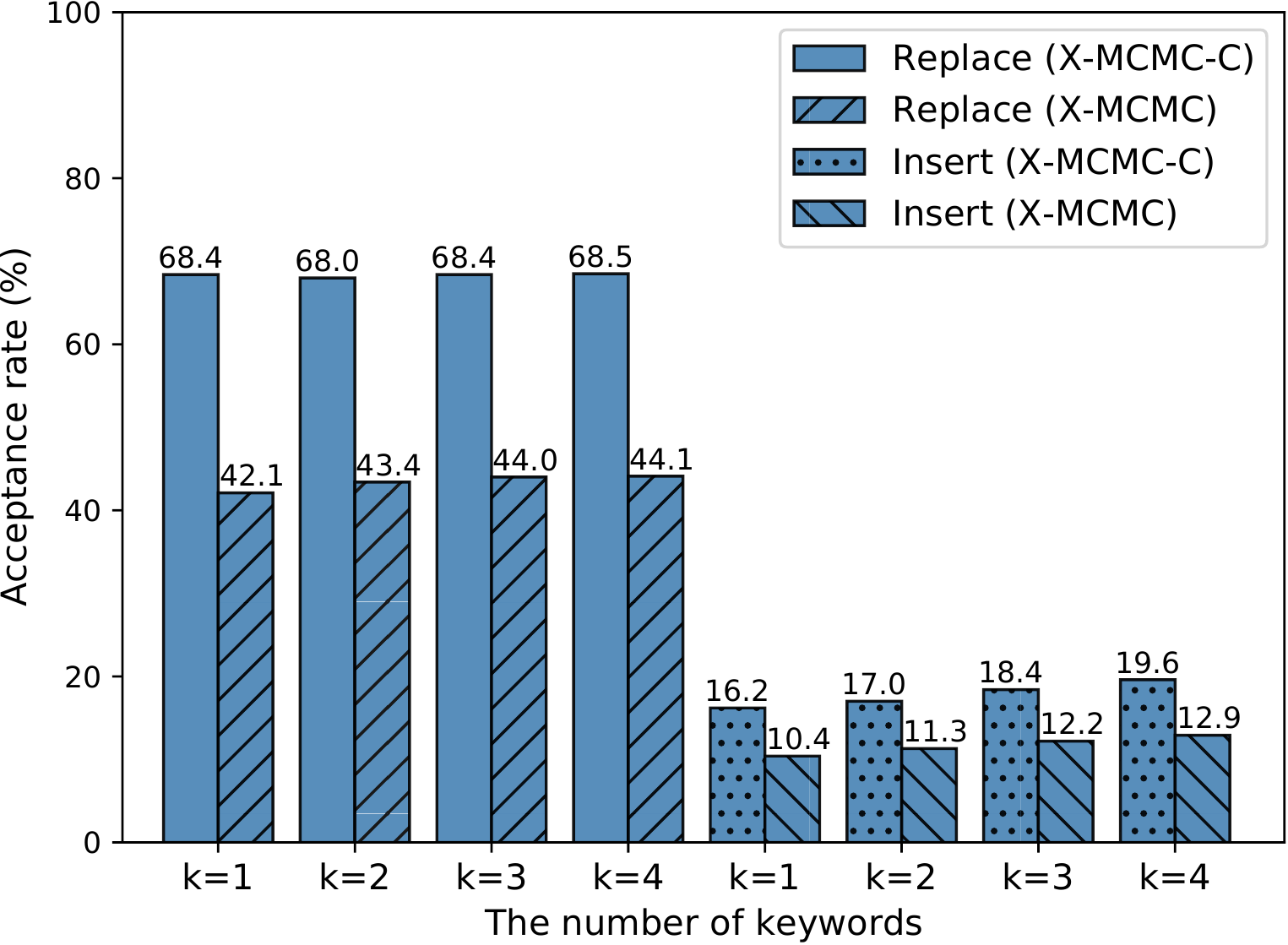} 
    \caption{Acceptance rates vs. the number of constraints $k$.}
    \label{acceptance2}
\end{figure}

We show the acceptance rates of XLNet-based models (X-MCMC and X-MCMC-C) in Figure \ref{acceptance2}.
Compared with X-MCMC, X-MCMC-C has much higher acceptance rates, consistent with their
LSTM-based counterparts. 

\section{Sentences Generated with Lexical Constraints}\label{appendix_c}
We show some sentences generated by different models with lexical constraints in Table \ref{tab:case2}. 
All MCMC-based models (CGMH, L-MCMC-C, and X-MCMC-C) are run for 200 steps. 
\begin{table*}[h]
  \centering
  \begin{tabular}{
    m{0.11\textwidth}<{\centering}|
    m{0.83\textwidth}<{}
    }

    \toprule
    \textbf{Constraints} & \textbf{view}\\
    \midrule
    Human&Victims commissioner Brendan McAllister has asked the Policing Board for its \textbf{view} on the transfer .\\
    \midrule
    sep-B/F& He said : `` There is a point of \textbf{view} , " he said in a statement . \\
    \midrule
    asyn-B/F& He said : `` There is a point of \textbf{view} , " he told BBC Radio 4 's Today programme . \\
    \midrule
    GBS & `` It 's a good thing to do , " he said , adding that he did not have a \textbf{view} of the situation .\\
    \midrule
    CGMH& If this case ends up or next to the \textbf{view}\\
    \midrule
    \textbf{L-MCMC-C} & What is the world 's \textbf{view} of the global economy ?\\
    \midrule
    \textbf{X-MCMC-C}& The consensus \textbf{view} is that this has not been achieved .\\

    \midrule
    \midrule
    \textbf{Constraints} & \textbf{promoting}, \textbf{energy}\\
    \midrule
    Human&They ranged from \textbf{promoting} clean \textbf{energy} and overseas infrastructure projects to attracting medical tourists and foreign firms .\\
    \midrule
    GBS &`` We are \textbf{promoting} \textbf{energy} efficiency , " said $<$UNK$>$ $<$UNK$>$ , a professor of economics at the University of California , Berkeley .\\
    \midrule
    CGMH& But it is now widely criticized for \textbf{promoting} the \textbf{energy}\\
    \midrule
    \textbf{L-MCMC-C}& He is also committed to \textbf{promoting} innovation in renewable \textbf{energy} projects .\\
    \midrule
    \textbf{X-MCMC-C}& They are expected to talk about curbing carbon emissions and \textbf{promoting} \textbf{energy} efficiency .\\

    \midrule
    \midrule
    \textbf{Constraints} & \textbf{experience}, \textbf{feels}, \textbf{dream}\\
    \midrule
    Human&`` This whole \textbf{experience}  \textbf{feels} like a \textbf{dream} come true . ...\\
    \midrule
    GBS & `` It \textbf{feels} like it 's a \textbf{dream} , " he said , adding that he had no \textbf{experience} in the world 's history .\\
    \midrule
    CGMH& In his \textbf{experience} , he \textbf{feels} he can 't find his own \textbf{dream}\\
    \midrule
    \textbf{L-MCMC-C} & The whole \textbf{experience} \textbf{feels} like a very different story -- a man 's \textbf{dream} .\\
    \midrule
    \textbf{X-MCMC-C}& The whole \textbf{experience} \textbf{feels} like a \textbf{dream} to me .\\
    \midrule
    \midrule
    \textbf{Constraints} & \textbf{seized}, \textbf{movement}, \textbf{party}, \textbf{taken}\\
    \midrule
    Human&They have \textbf{seized} on a common claim : A dangerous fringe \textbf{movement} , the tea \textbf{party} , has \textbf{taken} over the Republican Party .\\
    \midrule
    GBS & The \textbf{party} has \textbf{taken} part in a series of meetings with the opposition \textbf{movement} , which has been \textbf{seized} in recent years .\\
    \midrule
    CGMH& He has also \textbf{seized} his head off his \textbf{movement} toward his own \textbf{party} and has \textbf{taken}\\
    \midrule
    \textbf{L-MCMC-C} & The MDC has \textbf{seized} power , and the Islamist \textbf{movement} 's main opposition \textbf{party} has been \textbf{taken} over .\\
    \midrule
    \textbf{X-MCMC-C}& The two \textbf{seized} rebel \textbf{movement} leaders and the local \textbf{party} leader were also \textbf{taken} to police stations .\\
    \midrule
    \midrule
    \textbf{Constraints} & \textbf{talking}, \textbf{something}, \textbf{analysts}, \textbf{planning}\\
    \midrule
    Human&But it is unlikely he 'll be \textbf{talking} to victims , \textbf{something} Vatican \textbf{analysts} had suggested was a possibility in the early \textbf{planning} stages of the U.S. trip .\\
    \midrule
    GBS &`` I think it 's \textbf{something} we 're \textbf{talking} about , " said $<$UNK$>$ , who is \textbf{planning} to meet \textbf{analysts} on Wednesday .\\
    \midrule
    CGMH& After \textbf{talking} for \textbf{something} that happened , some \textbf{analysts} think it is \textbf{planning}\\
    \midrule
    \textbf{L-MCMC-C} & The first meeting came on Wednesday , after \textbf{talking} about \textbf{something} that some \textbf{analysts} were \textbf{planning} .\\
    \midrule
    \textbf{X-MCMC-C}& They are \textbf{talking} about \textbf{something} that some \textbf{analysts} have been \textbf{planning} for for several years .\\
    \bottomrule

  \end{tabular}
  \caption{Sentences generated by different
  models with lexical constraints on the One-Billion-Word test sets. (“Human” refers to the human reference.)}\label{tab:case2}
\end{table*}

\section{Text Infilling}\label{appendix_d}
We show some sentences generated by different models with different templates in Table \ref{tab:case3}.
\begin{table*}[h]
  \centering
  \begin{tabular}{
    m{0.18\textwidth}<{\centering}|
    m{0.72\textwidth}<{}
    }
    \toprule
    Template & Moving \underline{\qquad} from \underline{\qquad} induce problems \underline{\qquad} depression \underline{\qquad} \\
    \midrule
    Human Reference &Moving \textbf{them} from \textbf{person to person , home to home , can} induce problems \textbf{such as anxiety and} depression \textbf{.}\\
    \midrule
    FLM& Moving \textbf{forward} from \textbf{the end of the year , he said that} induce problems \textbf{could be used for} depression \textbf{.}\\
    \midrule
    BLM& Moving \textbf{away} from \textbf{embryonic stem cells , which can be used to} induce problems \textbf{, heart disease and} depression \textbf{.}\\
    \midrule
    Bayesian MCMC& Moving \textbf{down} from \textbf{the top that you say is that they can} induce problems \textbf{as a result of} depression \textbf{.}\\
    \midrule
    TIGS (strategy 1)&Moving \textbf{away} from \textbf{actor David James Jones to work with music to} induce problems \textbf{in modern film classic} depression \textbf{shows}\\
    \midrule
    TIGS (strategy 2) & Moving \textbf{away} from \textbf{the United States , the United States has been} induce problems \textbf{in the past major} depression \textbf{.} \\
    \midrule
    \textbf{L-MCMC} & Moving \textbf{away} from \textbf{the health care system that does not} induce problems \textbf{can cause a serious} depression \textbf{.} \\
    \midrule
    \textbf{L-MCMC-C} & Moving \textbf{away} from \textbf{the U.S. banking system can also be used to} induce problems \textbf{related to global economic} depression \textbf{.}\\
    \midrule

    \midrule
    Template & Opponents \underline{\qquad} say U.S. \underline{\qquad} suffer under the climate bill \underline{\qquad} trade \underline{\qquad } changes \underline{\qquad} \\
    \midrule
    Human Reference &Opponents \textbf{of the tariff} say U.S. \textbf{manufacturing would} suffer under the climate bill \textbf{regardless of} trade \textbf{policy} changes \textbf{.}	\\
    \midrule
    FLM& Opponents \textbf{of the bill} say U.S. \textbf{banks are} suffer under the climate bill \textbf{. $<$EOS$>$} trade \textbf{and} changes \textbf{to}\\
    \midrule
    BLM& Opponents \textbf{are expected to} say U.S. \textbf{does not} suffer under the climate bill \textbf{, to} trade \textbf{the} changes \textbf{.}\\
    \midrule
    Bayesian MCMC& Opponents \textbf{are expected to} say U.S. \textbf{lawmakers might} suffer under the climate bill \textbf{in the} trade \textbf{of} changes \textbf{.}\\
    \midrule
    TIGS (strategy 1)& Opponents \textbf{Michael Jackson supporters} say U.S. \textbf{car owners} suffer under the climate bill \textbf{plan to} trade \textbf{the} changes \textbf{government}\\
    \midrule
    TIGS (strategy 2) & Opponents \textbf{of the bill} say U.S. \textbf{banks will} suffer under the climate bill \textbf{, which} trade \textbf{on} changes \textbf{in} \\
    \midrule
    \textbf{L-MCMC} & Opponents say U.S. \textbf{citizens will} suffer under the climate bill \textbf{and also} trade \textbf{the} changes \textbf{.} \\
    \midrule
    \textbf{L-MCMC-C} & Opponents say U.S. \textbf{companies will} suffer under the climate bill \textbf{and make} trade changes \textbf{.}\\
    \midrule

    \midrule
    Template & The doctor said \underline{\qquad} the newspaper reported .\\
    \midrule
    Human Reference &The doctor said \textbf{it was unclear whether she will walk again , even though some people can recover from such injuries ,} the newspaper reported .\\
    \midrule
    FLM& The doctor said \textbf{there was no evidence that he had been involved in the incident . $<$EOS$>$ . $<$EOS$>$ . $<$EOS$>$ ,} the newspaper reported .\\
    \midrule
    BLM& The doctor said \textbf{the company posted a net profit of more than \$ 1 billion by the end of last year ,} the newspaper reported .\\
    \midrule
    Bayesian MCMC& The doctor said \textbf{that a 21-year-old man whose body was not taken to the hospital would have suffered from stab wounds ,} the newspaper reported .\\
    \midrule
    TIGS (strategy 1)&The doctor said \textbf{two clothing company Medical Center 's National Park Service found pictures of pop star Michael Jackson Ray Allen 18} the newspaper reported .\\
    \midrule
    TIGS (strategy 2) & The doctor said \textbf{the boy was not a man who was not involved in the incident . $<$EOS$>$ , with the statement} the newspaper reported . \\
    \midrule
    \textbf{L-MCMC} & The doctor said \textbf{he was not too worried ,} the newspaper reported . \\
    \midrule
    \textbf{L-MCMC-C} & The doctor said \textbf{it was a possible suicide attack , but it was not immediately clear what happened at the time ,} the newspaper reported .\\
    \bottomrule
  \end{tabular}
  \caption{Example outputs of different models with different templates on the One-Billion-Word test sets.}\label{tab:case3}
\end{table*}

\end{document}